\documentclass{article}

\usepackage{microtype}
\usepackage{graphicx}
\usepackage{subcaption}
\usepackage{booktabs} 
\usepackage{comment}

\usepackage{hyperref}

\usepackage[preprint]{icml2026}

\usepackage{bm}
\usepackage{amsmath}
\usepackage{booktabs}
\usepackage[table,x11names,dvipsnames]{xcolor}
\usepackage{graphicx}
\usepackage{tikzducks}
\usepackage{adjustbox}
\usepackage{duckuments}
\usepackage[abs]{overpic}
\usepackage{nicefrac}
\usepackage{wrapfig}
\usepackage{soul}
\usepackage{amssymb}
\usepackage[table]{xcolor}
\usepackage{tabularray}
\UseTblrLibrary{booktabs}
\usepackage{bm}
\usepackage{booktabs}
\usepackage{multirow}
\usepackage{array}
\colorlet{oursgray}{gray!15}
\newcolumntype{C}[1]{>{\centering\arraybackslash}p{#1}}

\definecolor{turquoise}{cmyk}{0.65,0,0.1,0.1}
\definecolor{purple}{rgb}{0.65,0,0.65}
\definecolor{darkgreen}{rgb}{0.0, 0.5, 0.0}
\definecolor{darkred}{rgb}{0.5, 0.0, 0.0}
\definecolor{darkblue}{rgb}{0.0, 0.0, 0.5}
\definecolor{blue}{rgb}{0.0, 0.0, 1.0}
\definecolor{orange}{rgb}{1.0, 0.5, 0.0}
\definecolor{red}{rgb}{1.0, 0.0, 0.0}
\definecolor{cherry}{RGB}{186,12,47}
\definecolor{pink}{RGB}{117,107,177}
\definecolor{oursgray}{gray}{0.90}

\newcommand{\refsec}[1] {Sec.~\ref{#1}}
\newcommand{\refeq}[1] {Eq.~\eqref{#1}}
\newcommand{\reffig}[1] {Fig.~\ref{#1}}

\newcommand{\reftab}[1] {Table~\ref{#1}}
\newcommand{\refapx}[1] {App.~\ref{#1}}

\usepackage[textsize=tiny]{todonotes}

\icmltitlerunning{Factorized Neural Implicit DMD for Parametric Dynamics}

\begin{document}

\twocolumn[
  \icmltitle{Factorized Neural Implicit DMD for Parametric Dynamics}



  \icmlsetsymbol{equal}{*}

  \begin{icmlauthorlist}
    \icmlauthor{Siyuan Chen}{equal,ubc}
    \icmlauthor{Zhecheng Wang}{equal,uoft}
    \icmlauthor{Yixin Chen}{uoft}
    \icmlauthor{Yue Chang}{uoft}
    \icmlauthor{Peter Yichen Chen}{ubc}
    \icmlauthor{Eitan Grinspun}{uoft}
    \icmlauthor{Jonathan Panuelos}{uoft}
  \end{icmlauthorlist}

  \icmlaffiliation{uoft}{Department of Computer Science, University of Toronto, Toronto, Canada}
  \icmlaffiliation{ubc}{Department of Computer Science, University of British Columbia, Vancouver, Canada}

  \icmlcorrespondingauthor{Zhecheng Wang}{zhecheng@cs.toronto.edu}

  \icmlkeywords{Koopman operator, Dynamic Mode Decomposition, Neural Field}

  \vskip 0.3in
]



\printAffiliationsAndNotice{}  

\begin{abstract}
A data-driven, model-free approach to modeling the temporal evolution of physical systems mitigates the need for explicit knowledge of the governing equations. Even when physical priors such as partial differential equations are available, such systems often reside in high-dimensional state spaces and exhibit nonlinear dynamics, making traditional numerical solvers computationally expensive and ill-suited for real-time analysis and control. Consider the problem of learning a parametric flow of a dynamical system: with an initial field and a set of physical parameters, we aim to predict the system's evolution over time in a way that supports long-horizon rollouts, generalization to unseen parameters, and spectral analysis.

We propose a \textit{physics-coded} neural field parameterization of the Koopman operator's spectral decomposition. Unlike a physics-constrained neural field~\cite{raissi2019physics}, which fits a single solution surface, and neural operators~\cite{li2020fourier,lu2021learning}, which directly approximate the solution operator at fixed time horizons, our model learns a \textit{factorized} flow operator that decouples spatial modes and temporal evolution. This structure exposes underlying eigenvalues, modes, and stability of the underlying physical process to enable stable long-term rollouts, interpolation across parameter spaces, and spectral analysis. We demonstrate the efficacy of our method on a range of dynamics problems, showcasing its ability to accurately predict complex spatiotemporal phenomena while providing insights into the system's dynamic behavior.
\end{abstract}

\section{Introduction}
\label{sec:introduction}
Accurate and efficient prediction of parametric partial differential equations (PDEs) with machine learning techniques has gained significant attention. In many practical settings, the governing dynamics depend on a variety of physical parameters, including geometric descriptors, boundary conditions, initial conditions, and flow parameters such as viscosity. Learning models that can capture both the underlying dynamics and their dependence on such parameters remains challenging, particularly when long-term stability, interpretability, and generalization across unseen parameter configurations are required.

Recent advances in \emph{Physics-informed Neural Network}~\cite{raissi2019physics} and \emph{Neural Operators}~\cite{li2020fourier,lu2021learning} have demonstrated strong approximation capability for PDE solution operators. However, these approaches model the learned dynamics as black-box mappings over fixed temporal sequences, offering limited interpretability and often suffering from error accumulation or instability under long-term rollout. As an alternative, Koopman operator theory~\cite{koopman1931hamiltonian} formulates nonlinear dynamical systems through linear evolution in an appropriately lifted space, enabling spectral analysis, modal decomposition, and stable time integration. While Koopman-based neural operators~\cite{xiong2024koopman} have recently shown promise, they typically rely on a high-dimensional latent operator that is implicitly defined and lacks explicit physical or spectral structure, limiting generalization across varying parameter settings and geometric configurations.

In this work, we propose a \textit{physics-coded} neural Koopman framework for learning parametric PDE dynamics with explicit spectral structure and robust long-term prediction. Specifically, we adopt a reduced-order modeling perspective and characterize the system evolution with a parameterized DMD operator, whose eigenfunctions ($\Phi$) and eigenvalues ($\Lambda$) are modeled by neural fields conditioned on physical parameters. This formulation yields a compact, low-rank representation of the dynamics that supports efficient dynamic modeling for unseen PDE instances within the same family, and enables continuous-space evaluation. We learn Koopman eigenfunctions as \textit{complex-conjugate} pairs by jointly estimating their real and imaginary parts to form a single mode. Each learned pair is then fixed, and subsequent eigenfunctions are trained sequentially with \textit{orthogonality} enforced against all previously learned modes, yielding a well-conditioned spectral decomposition and improved interpretability in terms of coherent spatial modes and their associated frequencies and growth/decay rates.


We evaluate our method on a diverse set of parametric PDE benchmarks, including Burgers’ equation with varying viscosities, two-dimensional Navier–Stokes flow in the double shear layer configuration, the K\'arm\'an vortex street past a circular obstacle, and flow over parameterized airfoil geometries with varying shape and inflow orientation. Across all tasks, our model achieves accurate long-term prediction, improved stability, and strong generalization compared with representative neural operators, autoregressive, and Koopman-based baselines. Moreover, the learned modes exhibit clear physical interpretability, capturing coherent flow structures and dominant dynamical patterns across different parameter settings and geometric configurations.

In summary, our main contributions are:
\begin{itemize}
  \item a physics-coded neural Koopman framework for generalization to unseen PDE instances and robust long-term prediction;
  
  \item neural field parameterizations of $\Phi$ and $\Lambda$ conditioned on code parameters, enabling compact low-rank dynamics and continuous-space evaluation;
  
  \item a sequential learning strategy for complex-conjugate modes with orthogonality constraints, improving stability, conditioning, and interpretability.
\end{itemize}

\begin{figure}[ht]
    \centering
    \includegraphics[width=\linewidth]{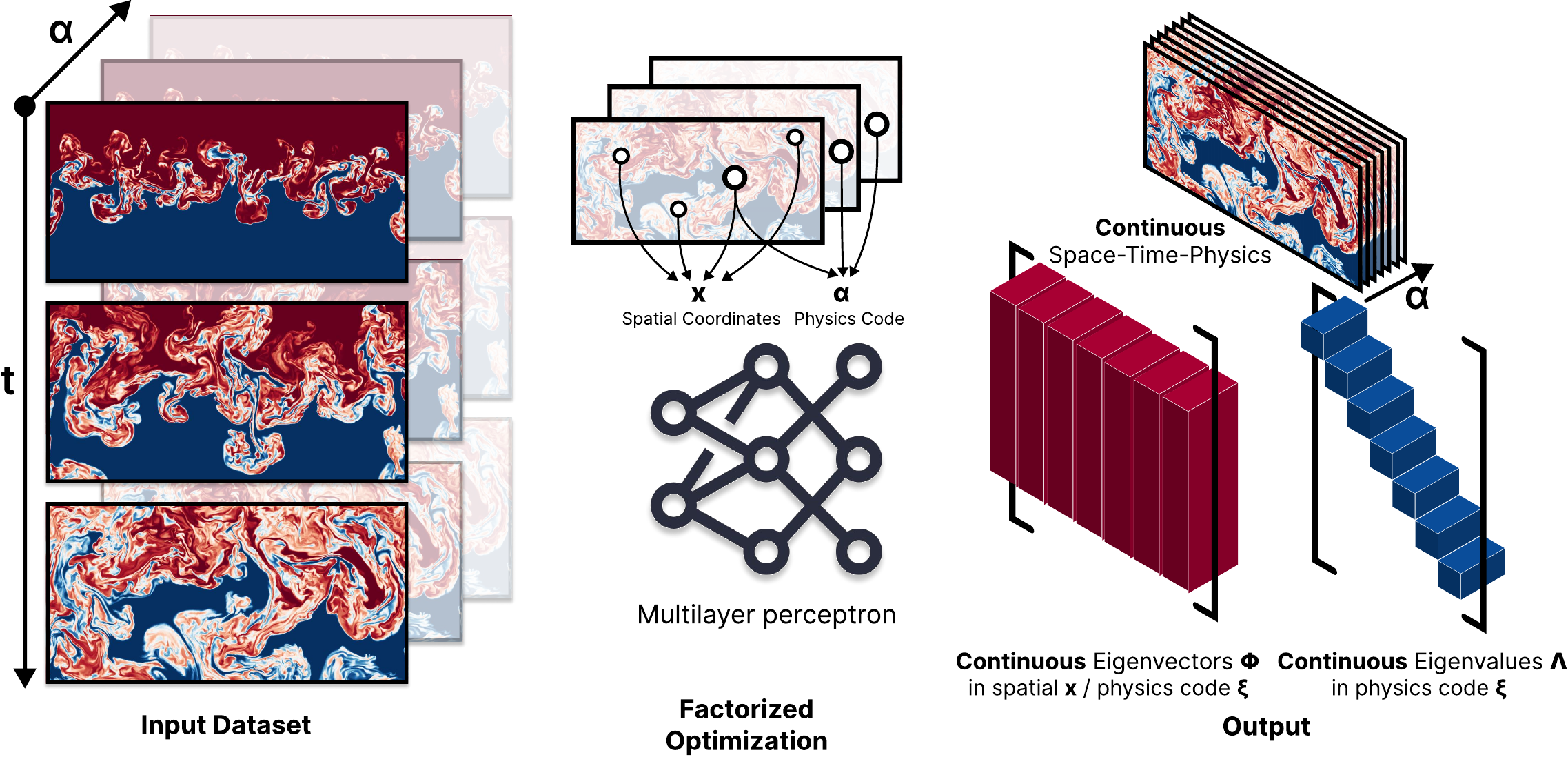}
    \caption{Physics-Coded Neural DMD Pipeline. From left to right: 
    (1) Input dataset at full resolution,  
    (2) Nonlinear optimization using spatial coordinates and physics code, 
    (3) Neural DMD output with continuous eigenvectors $\Phi$ and eigenvalues $\Lambda$ enabling space--time--physics reconstruction.}
    \label{fig:pipeline}
\end{figure}

\section{Related Works}
\label{sec:related_works}
In this section, we review prior work on data-driven modeling of PDE dynamics, neural operator learning, and autoregressive spatio-temporal prediction under the context of reduced-order spectral representations. We analyze their limitations and position our formulation in relation to these methods, highlighting how our proposed method addresses the resulting gaps.

\subsection{Koopman and Linear Latent Dynamics for PDEs}
\emph{Koopman theory}~\cite{koopman1931hamiltonian} provides a principled framework for capturing nonlinear dynamics through a globally linear representation, enabling modal decomposition and reduced-order modeling of complex systems like PDEs and fluid flows~\cite{Analysis2013,ROWLEY2009,Spectralanalysis2020}.

\emph{Dynamic Mode Decomposition (DMD)} offers a practical, data-driven approximation of the Koopman operator by identifying a linear propagator from temporal snapshots. Originally developed for engineering modal analysis~\cite{schmid2010dynamic}, DMD has been leveraged for efficient prediction in high-DoF simulations~\cite{chen2025fast}. To enhance its fidelity, extensions such as OptDMD~\cite{askham2018variable} and BOP-DMD~\cite{sashidhar2022bagging} incorporate advanced optimization and ensemble techniques, while continuous-time variants broaden the approach to nonlinear domains~\cite{rosenfeld2022dynamic}.

\emph{Koopman Autoencoders} extend this paradigm by learning latent spaces where evolution becomes approximately linear, improving long-term forecasting stability and interpretability~\cite{lusch2018deep}. Subsequent developments enforce bidirectional consistency, invertibility, and spectral stabilization to better preserve physical structure~\cite{azencot2020forecasting,meng2024koopman,choi2024koopman}. Separately, recent methods like ResKoopNet~\cite{xu2025reskoopnetlearningkoopmanrepresentations} aim to improve spectral approximation by explicitly minimizing the spectral residual to compute Koopman eigenpairs, addressing limitations in identifying the operator's complete spectrum. These advances connect classical Koopman theory to modern machine learning, as comprehensively reviewed by Brunton et al.~\cite{brunton2021modern}.

\subsection{Neural operators and PDE surrogates}
\emph{Neural operators} provide an alternative framework for modeling dynamics by learning mesh-agnostic mappings between function spaces, rather than finite-dimensional vectors. Prominent architectures include DeepONet~\cite{lu2021learning} and the Fourier Neural Operator (FNO)~\cite{li2020fourier}, which can generalize to unseen initial conditions and discretizations.
The Koopman Neural Operator (KNO)~\cite{xiong2024koopman} directly bridges this paradigm with Koopman theory by integrating a latent linear evolution operator within the FNO architecture, aiming to combine the representation power of neural operators with the structured temporal dynamics of the Koopman framework.

\subsection{Autoregressive and sequence-to-sequence models for PDE}

Autoregressive models have long been used to represent PDE dynamics. Early work explored direct autoregressive formulations \cite{barsinai2019learning, greenfeld2019learning, hsieh2019learningneuralpdesolvers}. Subsequent approaches improved efficiency and stability through methods like temporal bundling \cite{brandstetter2023message} and error accumulation analysis with stabilization strategies \cite{mccabe2023stability}.

Recently, token-based sequence-to-sequence models, inspired by the success of autoregressive language models \cite{radford2018improvingLU}, have become effective for scientific modeling. Several works apply autoregressive transformers to PDE learning. For instance, \cite{serrano2025zebra} uses an in-context learning approach for parametric PDEs, while \cite{morel2025disco} infers operator parameters without an explicit decoder. \cite{koupaï2025enma} develops a structured generative neural operator for uncertainty quantification, and \cite{holzschuh2025pde} proposes a transformer that treats spatio-temporal states as tokens. These represent a rapidly growing subset of research exploring token-based formulations for PDE operators.

In contrast to these nonlinear autoregressive models, our approach conditions on previous states but retains a linear reduced basis and linear time integration. This design yields faster inference and is inherently agnostic to the time step size.
\section{Physics-Coded Continuous Parameterization of the Koopman Operator}

We start with a continuous spatio-temporal dynamical system of some state $u$ defined by,
\begin{equation}
\label{eq:continuous_system_cond}
    \partial_t u(\bm{x},t;\bm{\xi}) = f\left(u(\bm{x},t;\bm{\xi}), \bm{x}; \bm{\xi}\right)
\end{equation}
with spatial coordinates $\bm{x}\in\Omega\subset\mathbb{R}^d$, time $t\ge0$, and physics code $\bm{\xi} \in \mathbb{R}^m$.
Physics code represents a parametric encoding of the physical configuration (e.g., boundary conditions, initial conditions, latent variables) that remains fixed per sequence. \refeq{eq:continuous_system_cond} thus defines a family of autonomous dynamical systems parameterized by $\bm{\xi}$.

\subsection{Continuous, Conditioned Koopman Expansion}
Let $F^t_{\bm{\xi}}$ denotes the flow map induced by \refeq{eq:continuous_system_cond} subject to $\bm{\xi}$, i.e.
$u(\cdot,t;\bm{\xi}) = F^t_{\bm{\xi}}\!\left(u(\cdot,0;\bm{\xi})\right)$.
The $\bm{\xi}$-conditioned Koopman operator $\mathcal{K}^t_{\bm{\xi}}$ acts linearly on observables $g$ of the state via
\begin{equation}
(\mathcal{K}^t_{\bm{\xi}} g)(u) \;=\; g\!\left(F^t_{\bm{\xi}}(u)\right),
\label{eq:koopman_def}
\end{equation}
and admits eigenfunctions $\{\psi_i(\cdot;\bm{\xi})\}_{i\ge 1}$ with associated continuous-time eigenvalues $\omega_i(\bm{\xi})\in\mathbb{C}$ such that
\begin{equation}
\mathcal{K}^t_{\bm{\xi}} \psi_i(\cdot;\bm{\xi}) \;=\; e^{\omega_i(\bm{\xi})t}\,\psi_i(\cdot;\bm{\xi}).
\label{eq:koopman_eig}
\end{equation}
For the vector-valued observable $g(u)=u$, suppose that $g$ admits the Koopman expansion
\begin{equation}
g(u)(\bm{x}) = \sum_{i=1}^\infty \phi_i(\bm{x};\bm{\xi})\,\psi_i(u;\bm{\xi}),
\end{equation}
where $\phi_i$ are Koopman modes corresponding to their respective $\psi_i$.
Evaluating along trajectories yields
\begin{align}
u(\bm{x},t;\bm{\xi})
&= \sum_{i=1}^\infty z_i(t;\bm{\xi})\,\phi_i(\bm{x};\bm{\xi}),\label{eq:z_continuous_and_state} \\
z_i(t;\bm{\xi}) &:= \psi_i(u(\cdot,t;\bm{\xi});\bm{\xi}).
\end{align}
Here, $z_i(t;\bm{\xi})$ are now time-dependent mode coefficients corresponding to $\phi_i$. For a discrete sampling interval $\Delta t$, letting $\lambda_i(\bm{\xi}) := e^{\omega_i(\bm{\xi})\Delta t}$ gives the linear evolution:
\begin{align}
u(\bm{x},t+\Delta t;\bm{\xi}) &= \sum_{i=1}^{\infty} \lambda_i(\bm{\xi})\, z_i(t;\bm{\xi})\, \phi_i(\bm{x};\bm{\xi}), \\
z_i(t+\Delta t;\bm{\xi}) &= \lambda_i(\bm{\xi})\, z_i(t;\bm{\xi}). 
\label{eq:z_discrete_and_state}
\end{align}
Further details are provided in Appendix~\ref{app:koopman_operator}.

\subsection{Truncation and Time Discretization}

For computational tractability, we truncate \refeq{eq:z_continuous_and_state} into a finite dimension $r$, producing the approximate expansion:
\begin{equation}
    u(\bm{x},t;\bm{\xi}) \approx \bm{z}(t;\bm{\xi})^T \bm{\Phi}(\bm{x};\bm{\xi}),\label{eq:truncated_Koopman}
\end{equation}
defined by the finite-dimensional reduced state $\bm{z}(t;\bm{\xi})=\sum_{i=1}^r z_i(t;\bm{\xi})\, \bm{e}_i,$ and $\bm{\xi}$-conditioned basis functions $\bm{\Phi}(\bm{x};\bm{\xi})=\sum_{i=1}^r \phi_i(\bm{x};\bm{\xi})\, \bm{e}_i$.
Physically, such truncation captures the top $r$ most important modes, omitting smaller details. Truncating \refeq{eq:z_discrete_and_state} gives the linear evolution of the modal coefficients:
\begin{equation}
    \label{eq:discrete_modal_cond}
    \bm{z}(t+\Delta t;\bm{\xi}) = \bm{\Lambda}(\bm{\xi}) \, \bm{z}(t;\bm{\xi}),
\end{equation}
with a diagonal matrix of eigenvalues
$\bm{\Lambda}(\bm{\xi}) = \mathrm{diag}\left(\lambda_1(\bm{\xi}),\lambda_2(\bm{\xi}),\dots,\lambda_r(\bm{\xi})\right)$, $\lambda_i(\bm{\xi}) = e^{\omega_i(\bm{\xi}) \Delta t}$.

\subsection{Neural Field Parameterization of Physics Sequences}

To allow for generalization across sequences with different physical settings, we parameterize both the basis functions $\bm{\Phi}$ and the eigenvalue operator $\bm{\Lambda}$ using a neural field conditioned on both coordinates $\bm{x}$ and physics code $\bm{\xi}$.

We thus express the basis functions $\bm{\Phi}_\theta$ as a continuous neural field function $\bm{\Phi}_\theta(\bm{x};\bm{\xi}) = \sum_{i=1}^r \phi_i(\bm{x};\bm{\xi})\bm{e}_i$, where $\theta$ denotes learnable network parameters.
Similarly, another network $\bm{\Lambda}_\eta$ will parameterize the eigenvalues (vector) as 
$\bm{\Lambda}_\eta(\bm{\xi}) = \mathrm{diag}\left(\lambda_1(\bm{\xi}), \lambda_2(\bm{\xi}), \dots, \lambda_r(\bm{\xi})\right)$,
with network parameters $\eta$.

\paragraph{Conjugate-Pair Parameterization}
\label{sec:conjugate_pair}
Rather than explicitly parameterizing the full \(r \times r\) truncated Koopman matrix, we directly parameterize the temporal evolution of the modal coefficients using complex spectral parameters \(\omega_i(\bm{\xi})\), according to \refeq{eq:discrete_modal_cond}.

Given a complex modal eigenvalue \(\omega_i(\bm{\xi}) = \alpha_i(\bm{\xi}) + i \beta_i(\bm{\xi})\), the real term represents the growth/decay rate and the imaginary term represents the oscillatory frequency.
The field \(u\) then is composed of basis functions each corresponding to a specific modal frequency \(\beta_i\). These basis functions naturally come in \textit{conjugate pairs}; for each complex basis function \(\phi_p(\bm{x}; \bm{\xi})\), there exists a conjugate basis function \(\phi_{p+P}(\bm{x}; \bm{\xi}) = \overline{\phi_p(\bm{x}; \bm{\xi})}\), with the corresponding eigenvalues \(\omega_p(\bm{\xi}) = \alpha_p(\bm{\xi}) + i \beta_p(\bm{\xi})\) and \(\omega_{p+P}(\bm{\xi}) = \overline{\omega_p(\bm{\xi})}\), respectively. As the coefficient \(z_p(t)\) evolves over time, the state oscillates between the real and imaginary components of each conjugate pair, which we refer to as a \textit{mode}, reflecting the system's dynamic behavior.

Consequently, each primary mode is represented by two basis functions and corresponding eigenvalues:
\[
\phi_p = \phi_p^{(r)} + i \phi_p^{(i)}, \qquad \phi_{p+P} = \phi_p^{(r)} - i \phi_p^{(i)},
\]
\[
w_p = w_p^{(r)} + i w_p^{(i)}, w_{p+P} = w_p^{(r)} - i w_p^{(i)},
\]
Both $\bm{\Phi}_\theta$ and $\bm{\Lambda}_\eta$ are parameterized as two real channels: one for the real part and one for the imaginary part.

\paragraph{Prediction and Evolution:}
To predict future states from a given sequence, we first obtain the modal coefficients $\bm{z}(t_i;\bm{\xi})$ by using the least squares inverse:
\begin{align}
    \bm{\Phi}_\theta^{\dagger}(\bm{\xi}) = (\bm{\Phi}_\theta^H \bm{\Phi}_\theta)^{-1} \bm{\Phi}_\theta^H. \label{eq:pinv}
\end{align}
We can then project the initial full state $u(\bm{x},t_i;\bm{\xi})$ onto the learned basis $\bm{\Phi}_\theta(\bm{x};\bm{\xi})$ with $\bm{z}(t_i;\bm{\xi}) =  \bm{\Phi}_\theta^{\dagger}(\bm{\xi})u(t_i;\bm{\xi})$.
Once the modal coefficients are obtained, future states ($t_j>t_i$) are predicted using the matrix exponential of the Koopman operator:
\begin{equation}
\hat{u}(\bm{x},t_j;\bm{\xi}) = \bm{\Phi}_\theta(\bm{x};\bm{\xi}) \cdot \exp\left[\bm{\Lambda}_\eta(\bm{\xi})(t_j - t_i)\right] \cdot \bm{z}(t_i;\bm{\xi}).
\end{equation}

\subsection{Optimization of the Neural Field Model}
We optimize the network parameters $\theta$ and $\eta$ by combining a multi-step prediction term with the deflation technique. The optimization pipeline consists of two key components: (1) short/long horizon prediction, (2) peeling off modes.

\subsubsection{Short/Long Horizon Prediction}
\label{sec:short_long_horizon}
We train the networks $\bm{\Phi}_\theta$ and $\bm{\Lambda}_\eta$ to approximate continuous basis functions and eigenvalues. To ensure both local accuracy and stable long-term rollouts, we introduce a mixed objective that jointly enforces short-horizon prediction fidelity and long-horizon temporal consistency.

Concretely, we combine a one-step prediction loss in the original DMD formulation \cite{schmid2010dynamic}  with a multi-step OptDMD-style loss \cite{askham2018variable}. The total training objective is
\begin{equation}
\label{eq:koopman_mixed_loss}
\mathcal{L} = \alpha\underbrace{\mathcal{L}_{\text{short}}}_{\text{DMD}}\ +\ \beta\underbrace{\mathcal{L}_{\text{long}}}_{\text{OptDMD}},
\end{equation}
where $\alpha$ and $\beta$ are the per-term weights.

\paragraph{Short-Horizon Prediction}
The short-horizon loss enforces accurate one-step transitions between consecutive snapshots, analogous to standard DMD supervision:
\begin{equation}
\label{eq:koopman_short_loss}
\begin{aligned}
\mathcal{L}_{\text{short}}
&= \sum_{i=0}^{N-1}
\Big\|
u(\bm{x},t_{i+1};\bm{\xi})
 -
\bm{\Phi}_\theta \bm{\Lambda}_\eta
\bm{\Phi}_\theta^{\dagger}
\hat{u}(\bm{x},t_i;\bm{\xi})
\Big\|_2^2 .
\end{aligned}
\end{equation}

\paragraph{Long-Horizon Prediction}
Short-horizon supervision alone does not constrain accumulated errors over long rollouts. To prevent short-horizon overfitting, we additionally enforce multi-step consistency using repeated application of the Koopman operator, following an OptDMD-style formulation:
\begin{equation}
\label{eq:koopman_long_loss}
\mathcal{L}_{\text{long}}
= \sum_{j=1}^{N}
\Big\|
u(\bm{x},t_j;\bm{\xi}) -
\bm{\Phi}_\theta \bm{\Lambda}_\eta^{j}
\bm{\Phi}_\theta^{\dagger}
\hat{u}(\bm{x},t_0;\bm{\xi})
\Big\|_2^2 .
\end{equation}

\subsubsection{Stage-wise Deflation for Spectral Modes}
\label{sec:deflation}
Following \refsec{sec:conjugate_pair}, a trajectory can be decomposed into multiple spectral modes at different frequencies, where each mode is represented by a conjugate pair of basis functions and eigenvalues. In practice, however, learning all modes jointly often produces \emph{highly correlated} modes: many modes compete for the same low-frequency band, leading to redundant bases.

To promote clearer spectral separation, we encourage the learned basis functions corresponding to different modes to be orthogonal under \emph{conjugate orthogonality}. Naive orthogonality regularization often leads to poor conditioning and unstable optimization; \reffig{fig:ablation} shows that jointly learning all modes with such a regularizer results in noisy and highly correlated modes (e.g., the 2nd and 3rd modes). Instead, we adopt a stage-wise deflation strategy that enforces orthogonality by construction.

Suppose the first $(p-1)$ modes have already been learned. At stage $p$, we freeze all previously learned basis-function and eigenvalue networks, and perform the following steps:
\begin{enumerate}
    \item Evaluate the frozen networks to obtain $\bm{\Phi}_{<p}(\bm{x};\bm{\xi})$ and compute its pseudoinverse $\bm{\Phi}_{<p}^{\dagger}$.
    
    \item Evaluate the current networks to produce a raw candidate mode $\tilde{\bm{\phi}}_{p}(\bm{x};\bm{\xi})$ and its associated eigenvalue $\omega_p(\bm{\xi})$, with gradients enabled.
    
    \item Deflate the raw mode by removing its projection onto the span of previously learned modes:
    \begin{equation}
    \bm{\phi}_{p} \;\leftarrow\; \big(\bm{I}-\bm{\Phi}_{<p}\bm{\Phi}_{<p}^{\dagger}\big)\,\tilde{\bm{\phi}}_{p},
    \end{equation}
    where $\bm{\Phi}_{<p}$ and $\bm{\Phi}_{<p}^{\dagger}$ are treated as stop-gradient operators so that gradients flow only through $\tilde{\bm{\phi}}_{p}$.
\end{enumerate}

Finally, we append the conjugate counterpart $\phi_{p+P}=\overline{\phi_p}$ and eigenvalue $\omega_{p+P}(\bm{\xi})=\overline{\omega_p(\bm{\xi})}$, solve the modal coefficients via least squares, and minimize the prediction loss. Backpropagation updates only the stage-$p$ networks, yielding progressively de-correlated modes with less spectral overlap.
\section{Experiments}
\label{sec:results}

We evaluate the proposed \emph{neural implicit DMD operator} (denoted as \textbf{INR-DMD} where INR stands for \textit{Implicit Neural Representation}) on a diverse set of dynamical systems, covering physics-coded PDE families with varying physical conditions. 
Specifically, we report results on Burgers' equation with varying viscosity (\refsec{sec:burgers}), 2D Navier--Stokes vorticity dynamics (\refsec{sec:navier_stokes}), the K\'arm\'an vortex street with parametrized geometry and boundary conditions (\refsec{sec:karman}) and an inflow airfoil with a six-dimensional shape space (\refsec{sec:airfoil}). All datasets are generated using high-fidelity numerical solvers. All training and evaluation experiments were conducted on a single RTX 4090 GPU.

\paragraph{Baselines.}
Per discussion in \refsec{sec:related_works}, we compare against Consistent Koopman-AE (KAE)~\cite{azencot2020forecasting}, Fourier Neural Operator (FNO)~\cite{li2020fourier}, Koopman Neural Operator (KNO) ~\cite{xiong2024koopman}, Parametric DMD (P-DMD)~\cite{pDMD}, PDE-Transformer (PDE-T)~\cite{holzschuh2025pde} and ResKoopNet (RKN)~\cite{xu2025reskoopnetlearningkoopmanrepresentations}. 

\paragraph{Metrics and timing.}
Prediction accuracy is evaluated using the relative mean squared error (rMSE) between the predicted state fields and the ground truth at each time step, computed on a test set consisting of unseen code parameters. Inference efficiency is measured by the average prediction time per frame in milliseconds. Each number is rounded to three decimal places.

\subsection{Burgers' Equation}
\label{sec:burgers}

We study the 1D viscous Burgers' equation
\begin{equation}
\label{eq:burgers}
\begin{aligned}
\partial_t u(x,t) + \partial_x\!\left(\frac{u(x,t)^2}{2}\right)
&= \nu\,\partial_{xx}u(x,t), \\
x\in(0,1), t\in(0,1], &\qquad
u(x,0)=u_0(x),
\end{aligned}
\end{equation}
where the viscosity $\nu$ serves as the \emph{physics code}. We evaluate generalization across a family of PDEs by varying $\nu\in[0.001,\,0.01]$. During training, we uniformly sample 10 viscosity values and generate reference trajectories. For testing, we evaluate on 9 unseen viscosities. Further details on the dataset generation are provided in \refapx{app:burgers}. 
\paragraph{Results}
We outperform all baselines both qualitatively and quantitatively, as shwon in \reftab{tab:experiment}. Our method achieves the lowest error while using the third-smallest number of parameters (the smaller models, KNO and KAE, incur 100× slower inference time and 10× prediction error). As shown in \reffig{fig:burgers_full_physics}, only our method and P-DMD can accurately capture and simulate the variations induced by changing viscosity; moreover, compared with P-DMD, our approach achieves this with only $1/5$ of the parameters and a 60× speedup in inference time, demonstrating robust generalization within the physical parameter space.

\begin{figure}[t]
  \centering
  \includegraphics[width=\linewidth]{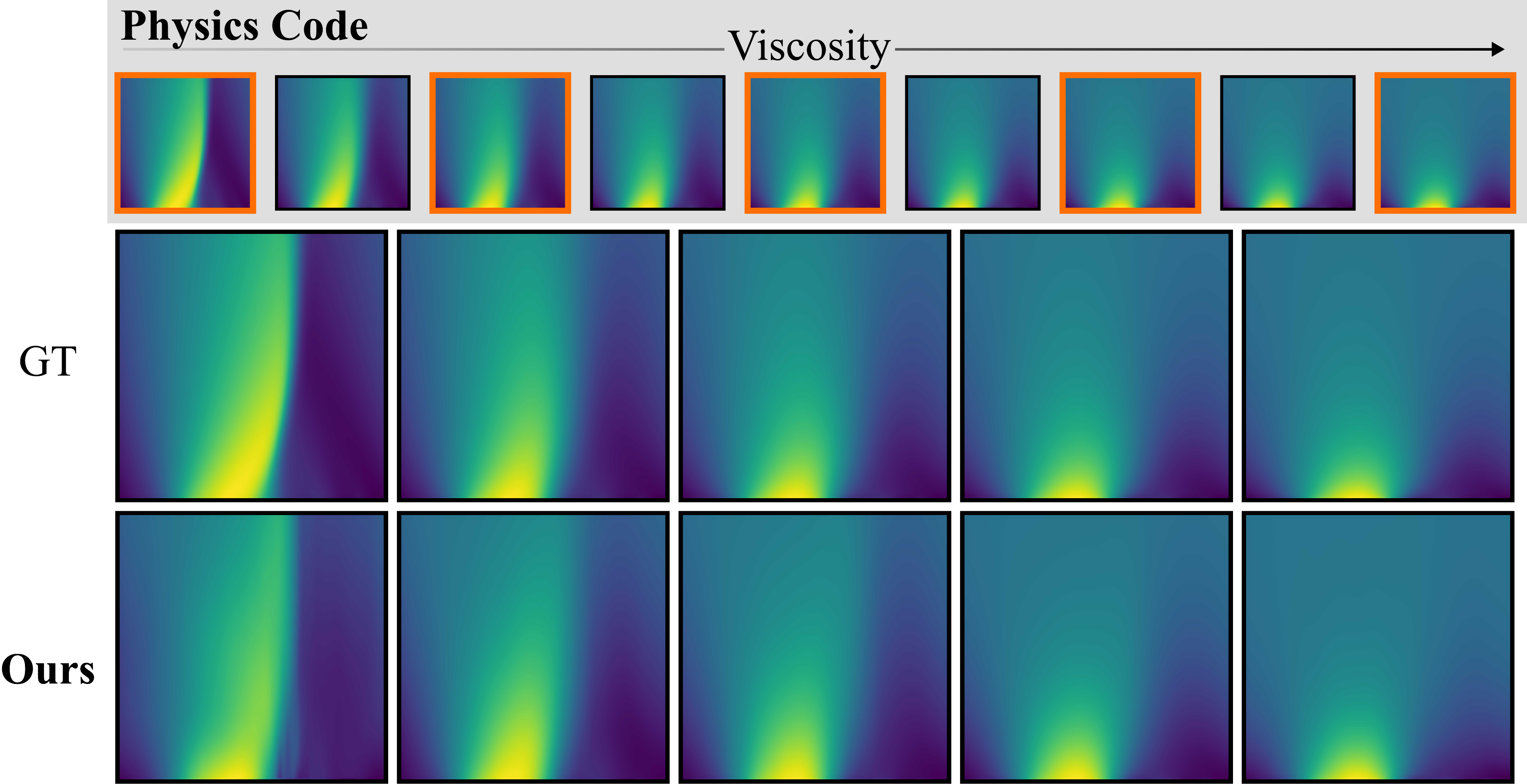}
  \caption{\textbf{Burgers' Equation.}
  Spatio-temporal visualization of the solution $u(x,t)$ for the same test case
  (horizontal axis: $x$, vertical axis: $t$, $u(0,0)$ corresponds to the bottom-left corner of each subplot.}
  \label{fig:burgers}
\end{figure}

\subsection{Double Shear Layer}
\label{sec:navier_stokes}
We consider the 2D incompressible Navier--Stokes equations in vorticity form
\begin{equation}
\label{eq:ns_vorticity}
\partial_t \omega + u\cdot\nabla\omega
= \nu\,\Delta\omega + f,
\end{equation}
with initial condition $\omega(\cdot,0) = \omega_0$. The initial condition is a \emph{double shear layer} that triggers Kelvin--Helmholtz instability, producing vortex roll-up and strong nonlinear interactions. We simulate for $t=8.0$, storing 160 snapshots. Each sequence differs by the initial shear-layer separation $s\in[0.2,0.4]$. We uniformly sample 11 values for training and randomly select 10 others for testing. See \refapx{app:double} for dataset generation details.

\paragraph{Results}
Our method achieves the lowest error while maintaining the fastest inference speed. Compared to methods with a comparable number of parameters (KAE and KNO), our approach is significantly faster and yields more than an order-of-magnitude reduction in error. Qualitatively, our method best captures the vortex structures arising from different initial conditions, demonstrating strong generalization across varying initial states.

\subsection{K\'arm\'an Vortex Street}
\label{sec:karman}

\begin{figure}
    \centering
    \includegraphics[width=\linewidth]{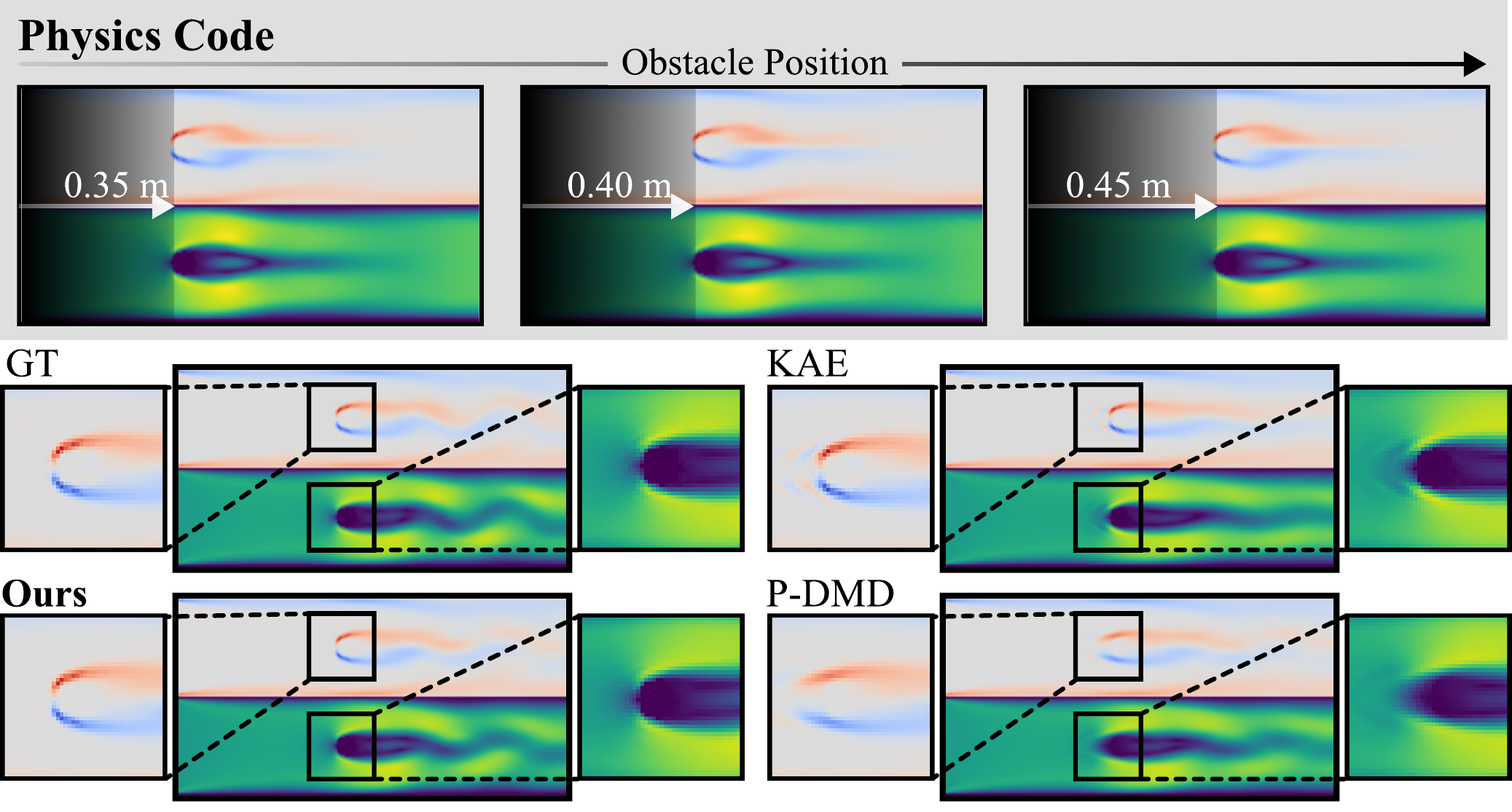}
    \caption{\textbf{K\'arm\'an vortex street.}
    Qualitative comparison of flow-field predictions under varying obstacle positions. KAE and P-DMD exhibit a noticeable double-shadow artifact (see zoom-ins), while our neural-field parameterization yields smoother transitions across physics conditions and remains close to the ground truth (GT). All heatmaps share the same color scale.}
    \label{fig:karman}
\end{figure}

We further evaluate generalization under \emph{parametrized boundary conditions} using the K\'arm\'an vortex street benchmark. All data are generated using the lattice Boltzmann method (LBM), of which the governing kinetic equation is the Boltzmann equation~\cite{Shan-2006}
\begin{equation}
\frac{\partial \bm{f}}{\partial t} + \bm{v}\cdot\nabla \bm{f} = \Omega(\bm{f}) + \bm{F}\cdot \nabla_{\bm{v}}\bm{f} \;,
\label{eq:boltzmann_equation}
\end{equation}
where $\bm{f}$ denotes the distribution function, $\bm{F}$ is external forces, and $\Omega$ is the collision operator that relaxes $\bm{f}$ towards the local equilibrium $\bm{f}^\text{eq}$~\cite{coreixas2017recursive}. The macroscopic quantities, such as density $\rho$ and velocity $\bm{u}$, can be recovered from the distribution $\bm{f}$.

Compared to the previous experiments on fixed-domain PDEs, this setting is substantially more challenging because it introduces \emph{fluid--solid coupling}: the flow evolution is explicitly conditioned on the presence and location of a solid obstacle, and changing the cylinder placement alters the boundary constraints and the resulting wake dynamics.
Here, the cylinder location serves as the physics code. We vary the obstacle x-coordinate within $[0.35, 0.45]$. We consider 21 distinct cylinder placements, randomly select 11 locations for training, and evaluate on the remaining 10 unseen locations for testing. All data sequences are simulated for 200 time steps.

\paragraph{Results}
In this example, our method outperforms all baselines across all evaluated metrics, including memory usage, inference time, and error. As shown in \reffig{fig:karman_full_time}, our approach accurately reconstructs the boundary conditions and the resulting Kármán vortex street for different initial positions of the circular obstacle, demonstrating generalization to boundary conditions with varying spatial configurations.

\subsection{Airfoil with Shape Space}  
\label{sec:airfoil}

\begin{figure}[h]
    \centering
    \includegraphics[width=\linewidth]{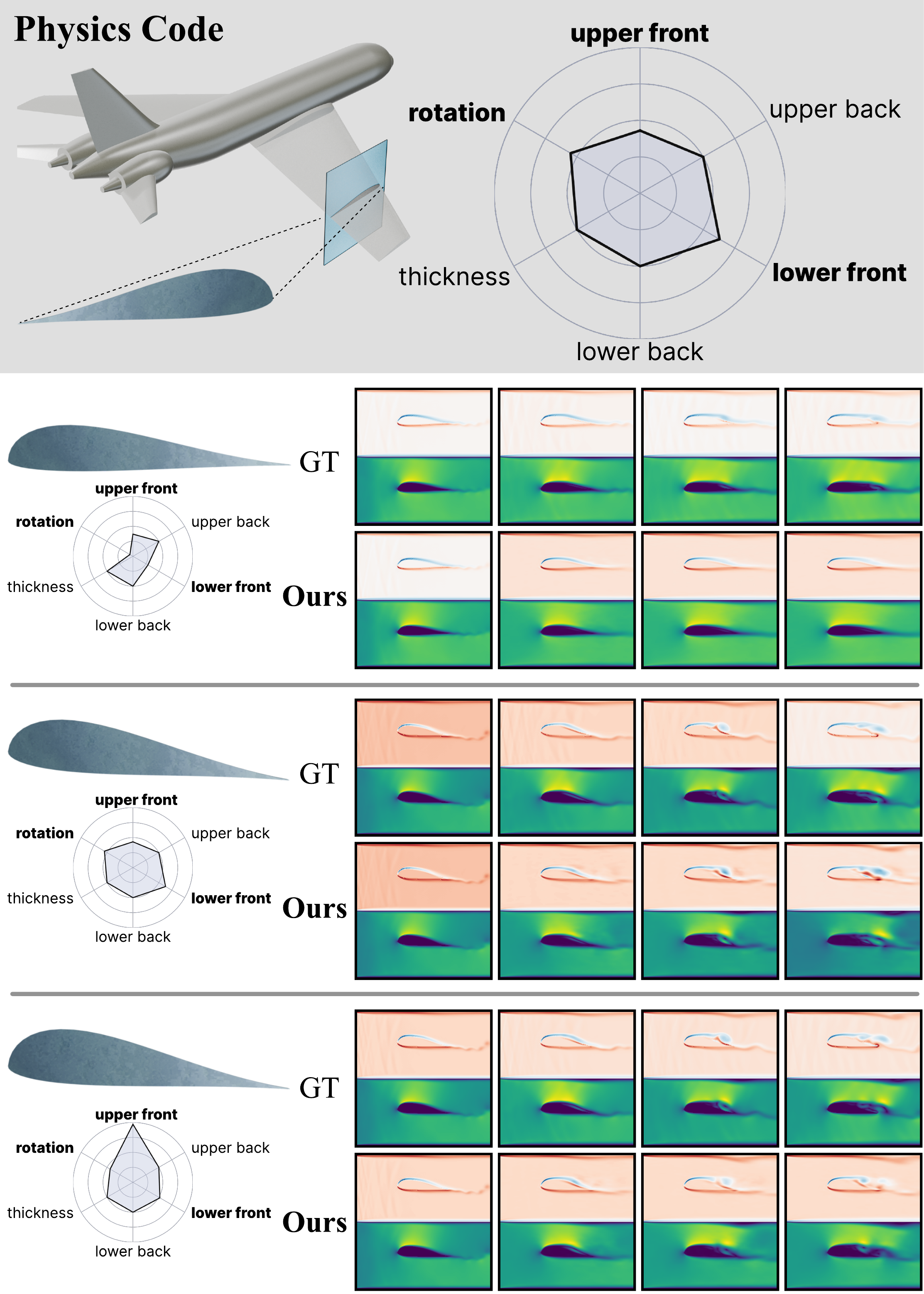}
    \caption{\textbf{Physics code for airfoil parameterization and qualitative rollouts.}
    \emph{Top:} Illustration of the physics code that maps a \textbf{6-dimensional} parameter vector to an airfoil geometry, including a radar-style view of the six CST parameters (upper/lower surface controls, trailing-edge thickness, and clockwise rotation). \emph{Bottom:} Representative examples showing the generated airfoil shapes (left) and the corresponding flow-field predictions over time (right), comparing ground truth (GT) with \emph{Ours} under different parameter settings in the test set. All heatmaps share the same color scale.}
    \label{fig:airfoil}
\end{figure}

Building upon the K\'arm\'an vortex street experiment, we explore a more challenging airfoil shape parameterized by six variables: $\{ A_{u0}, A_{u1}, A_{l0}, A_{l1}, t_e, \theta_{cw} \}$, where $A_{u0}, A_{u1}$ control the upper surface, $A_{l0}, A_{l1}$ the lower surface, $t_e$ the trailing-edge thickness, and $\theta_{cw}$ the clockwise rotation angle. This creates a six-dimensional parameter space with complex fluid--solid coupling. We sample 43 parameter points, varying $A_{u0}\in[0.40, 0.50]$, $A_{l0}\in[-0.20, -0.10]$, and $\theta_{cw}\in[0^\circ, 10^\circ]$. 27 points are used for training, 16 for testing. See \refsec{app:airfoil} for dataset details.

\paragraph{Results}
In this example, our method achieves the best performance across all evaluated metrics, including error, inference time, and memory consumption. As shown in \reffig{fig:airfoil}, our approach captures the distinct patterns induced by varying airfoil geometries and angles, demonstrating generalization along the shape (geometry) axis.

\subsection{Ablation Study}
\label{sec:ablation}

\begin{figure*}[ht]
    \centering
    \includegraphics[width=0.9\linewidth]{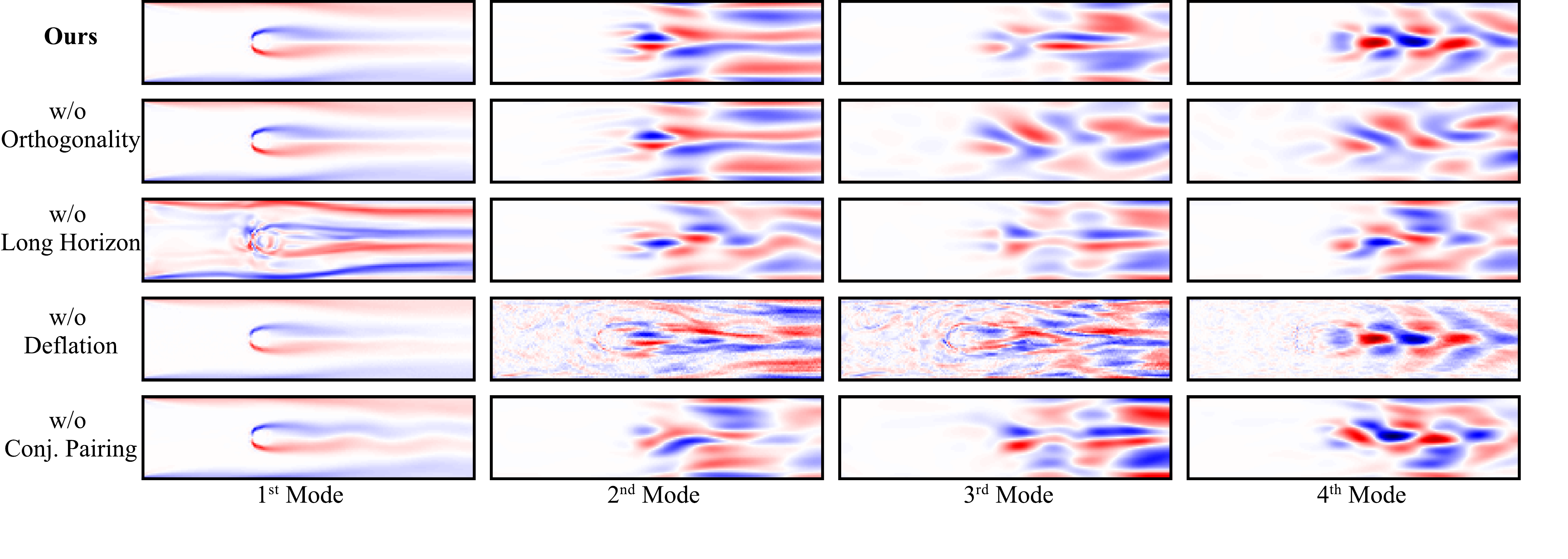}
    \caption{\textbf{Comparison of basis functions (real part) for the full INR-DMD and ablations.} 
    For the deflation baseline, we show the basis learned at each stage; for the non-deflation baseline, we visualize the four lowest-frequency modes. The full pipeline learns clean, symmetric, frequency-separated bases with low inter-mode correlation, whereas removing orthogonality, long-horizon loss, or deflation yields noisier and highly correlated modes, and removing conjugate pairing breaks spectral symmetry.
}
\label{fig:ablation}
\end{figure*}

\begin{figure}[!htp]
\centering
\includegraphics[width=0.9\linewidth]{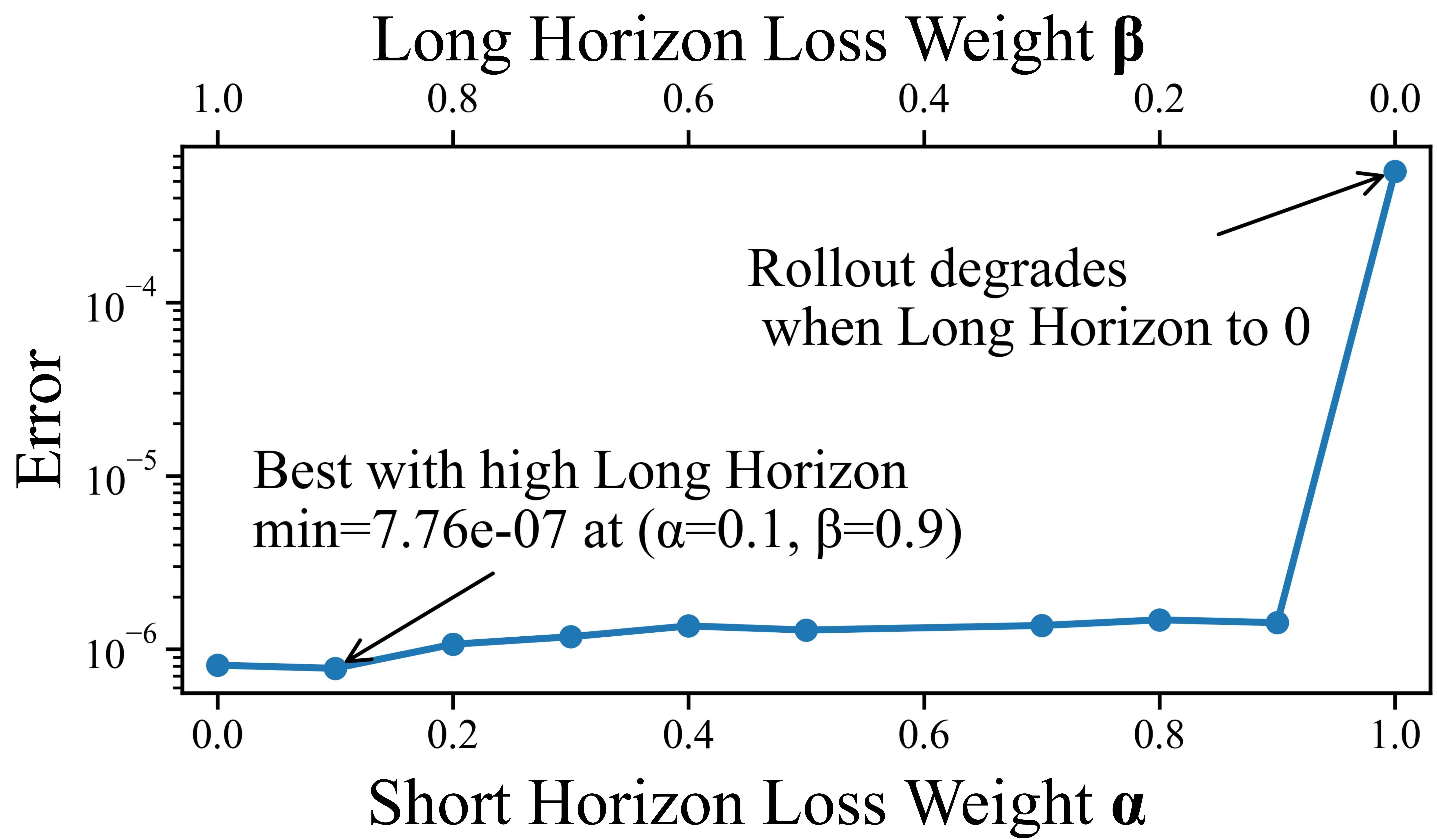}
\caption{\textbf{Short-horizon vs.\ long-horizon loss weighting sweep.}
We vary the prediction loss weights from $(\alpha,\beta)=(0,1)$ to $(1,0)$, where $\alpha$ and $\beta$ are short/long horizon loss weights respectively. The curve reports the resulting rollout error (lower is better) on the K\'arm\'an vortex street example.}
\label{fig:ablation_onestep_weight_sweep}
\end{figure}

Our full model comprises several coupled design choices for learning a finite-dimensional Koopman 
representation that enables robust long-horizon rollouts and interpretable spectra. We first sweep model size (varying the number of neurons and layers) to select a reasonable default capacity, and then ablate individual components one at a time. To isolate their individual contributions, we ablate one component at a time, keeping all other settings fixed (same latent rank $r$, training steps, optimizer, and identical train/test splits). We conduct ablations on the K\'arm\'an vortex street benchmark under geometry-conditioned settings (\refsec{sec:karman}).

\begin{table}[!hbp]
\centering
\begin{minipage}{\linewidth}
\centering
\caption{Ablation study of key design choices on K\'arm\'an vortex street. $\downarrow$: lower is better.}
\label{tab:ablation}
\renewcommand{\arraystretch}{1.15}
\begin{tabular}{c c c}
\toprule
\textbf{Variant} & \textbf{\# params} & \textbf{error $\downarrow$} \\
\midrule
\textbf{Ours} & 316,510 & $8.093\times10^{-7}$ \\
w/o Orthogonality & 316,510 & $1.156\times10^{-6}$ \\
w/o Long Horizon & 316,510 & $5.697\times10^{-4}$ \\
w/o Deflation & 354,172 & $1.819\times10^{-4}$ \\
w/o Conj. Pairing & 316,510 & $4.089\times10^{-6}$ \\
\bottomrule
\end{tabular}
\end{minipage}
\end{table}

\paragraph{w/o Orthogonality.}
With only deflation and conjugate pairing, the lack of orthogonality constraint causes the model to diverge over long horizons (\reffig{fig:ablation_onestep_weight_sweep}), increasing the rollout error from $8.093\times10^{-7}$ to $5.697\times10^{-4}$, and yields noisier, more correlated basis functions with noticeable mode mixing (\reffig{fig:ablation}).

\paragraph{w/o Long Horizon Loss.}
Removing long horizon loss leads to numerical instability over long horizons (\reffig{fig:ablation_onestep_weight_sweep}), increasing rollout error from $8.093\times10^{-7}$ to $5.697\times10^{-4}$ and producing noisier, more correlated bases with mode mixing compared to the smoother, more physically consistent modes learned with long horizon supervision (\reffig{fig:ablation}).

\paragraph{w/o Deflation.}
Training all modes jointly end-to-end with only a conjugate-orthogonality regularizer is insufficient, yielding markedly less conjugate-orthogonal and noisier modes (especially Modes~2--3; \reffig{fig:ablation}) and reducing rollout stability, with error increasing from $8.093\times10^{-7}$ to $4.089\times10^{-4}$ (\reftab{tab:ablation}).

\paragraph{w/o Conjugate Pairing.}
Removing the explicit conjugate-pairing constraint and learning $r$ complex modes independently produces a less symmetric spectrum (\reffig{fig:ablation}) and degrades long-horizon stability, increasing rollout error from $8.093\times10^{-7}$ to $4.089\times10^{-6}$.

\paragraph{Quantitative results.}
\reftab{tab:ablation} summarizes model parameter and rollout errors.
Across all benchmarks, the full INR-DMD consistently achieves the lowest rollout error, Notably, removing deflation yields the largest degradation in rollout errors.

\begin{table*}[!htbp] 
\centering
\caption{Results on multiple physics conditions.} 
\label{tab:experiment} 

\centerline{
\begin{minipage}{0.98\linewidth}
\centering
\small  
\begin{minipage}{0.49\linewidth}
\centering
\renewcommand{\arraystretch}{1.0}  
\begin{tabular}{@{}c@{\hspace{0.7em}}c@{\hspace{0.7em}}c@{\hspace{0.7em}}c@{\hspace{0.7em}}l@{\hspace{0.7em}}c@{\hspace{0.7em}}c@{}} 
\toprule 
\textbf{Exp.} & \textbf{\#P} & \textbf{\#S} & \textbf{Model}& \textbf{\#Params} & \textbf{Time} & \textbf{Error} \\ 
& & & & & \textbf{(ms)$\downarrow$} & \textbf{$\downarrow$} \\ 
\midrule 
\multirow{7}{*}{\shortstack{Burgers'\\Equation}} & \multirow{7}{*}{1} & \multirow{7}{*}{10} & KAE & 35,408& 1.269 & $2.35{\times}10^{-2}$ \\ 
& & & FNO & 287,425& 1.864 & $2.23{\times}10^{-2}$ \\ 
& & & KNO & 4,417& 2.577 & $5.70{\times}10^{-3}$ \\ 
& & & P-DMD & 253,575& 0.381 & $1.35{\times}10^{-3}$ \\ 
& & & PDE-T & N/A& N/A & N/A \\ 
& & & RKN & 5,820,249& 0.032 & $2.70{\times}10^{-3}$ \\ 
& & & \cellcolor{oursgray}INR-DMD & \cellcolor{oursgray}25,540& \cellcolor{oursgray}0.006 & \cellcolor{oursgray}$1.35{\times}10^{-4}$ \\ 
\midrule 
\multirow{7}{*}{\shortstack{Double\\Shear}} & \multirow{7}{*}{1} & \multirow{7}{*}{11} & KAE & 271,952& 1.278 & $4.78{\times}10^{-2}$ \\ 
& & & FNO & 1,188,353& 2.976 & DNF \\ 
& & & KNO & 263,297& 6.990 & $5.64{\times}10^{-2}$ \\ 
& & & P-DMD & 1,170,152& 5.445 & $5.24{\times}10^{-3}$ \\ 
& & & PDE-T & 33,190,328& 32.56 & $4.41{\times}10^{-2}$ \\ 
& & & RKN & 218,967,897& 1.827 & $7.62{\times}10^{-3}$ \\ 
& & & \cellcolor{oursgray}INR-DMD & \cellcolor{oursgray}378,264 & \cellcolor{oursgray}0.107 & \cellcolor{oursgray}$4.59{\times}10^{-3}$ \\ 
\bottomrule 
\end{tabular} 
\end{minipage}
\hfill
\begin{minipage}{0.49\linewidth}
\centering
\renewcommand{\arraystretch}{1.0}
\begin{tabular}{@{}c@{\hspace{0.7em}}c@{\hspace{0.7em}}c@{\hspace{0.7em}}c@{\hspace{0.7em}}l@{\hspace{0.7em}}c@{\hspace{0.7em}}c@{}} 
\toprule 
\textbf{Exp.} & \textbf{\#P} & \textbf{\#S} & \textbf{Model} & \textbf{\#Params} & \textbf{Time} & \textbf{Error} \\ 
& & & & & \textbf{(ms)$\downarrow$} & \textbf{$\downarrow$} \\ 
\midrule 
\multirow{7}{*}{\shortstack{K\'arm\'an\\Vortex}} & \multirow{7}{*}{1} & \multirow{7}{*}{11} & KAE & 678,192& 1.210 & $5.24{\times}10^{-6}$ \\ 
& & & FNO & 1,188,514& 2.600 & DNF \\ 
& & & KNO & N/A& N/A & N/A \\ 
& & & P-DMD & 2,436,392& 13.56 & $8.88{\times}10^{-7}$ \\ 
& & & PDE-T & N/A& N/A & N/A \\ 
& & & RKN & OOM& OOM & OOM \\ 
& & & \cellcolor{oursgray}INR-DMD & \cellcolor{oursgray}316,510& \cellcolor{oursgray}0.073 & \cellcolor{oursgray}$7.77{\times}10^{-7}$ \\ 
\midrule 
\multirow{7}{*}{\shortstack{Airfoil}} & \multirow{7}{*}{6} & \multirow{7}{*}{27} & KAE & 2,164,304& 1.288 & $7.74{\times}10^{-6}$ \\ 
& & & FNO & 1,188,514& 2.813 & $8.57{\times}10^{-6}$ \\ 
& & & KNO & N/A& N/A & N/A \\ 
& & & P-DMD & 6,181,752& 34.50 & $5.91{\times}10^{-6}$ \\ 
& & & PDE-T & 33,190,328& 33.06 & $8.67{\times}10^{-2}$ \\ 
& & & RKN & OOM& OOM & OOM \\ 
& & & \cellcolor{oursgray}INR-DMD & \cellcolor{oursgray}316,510& \cellcolor{oursgray}0.384 & \cellcolor{oursgray}$4.21{\times}10^{-6}$ \\ 
\bottomrule 
\end{tabular} 
\end{minipage}
\end{minipage}
}

\par\smallskip 
\noindent{\footnotesize \textit{Note:} $\downarrow$: lower is better; our method is in \colorbox{oursgray}{\strut gray}. \textbf{\#P}: physics condition dimension; \textbf{\#S}: number of trajectories in training set; N/A: failed to train; DNF: numerical overflow; OOM: out of memory.}
\end{table*}
\section{Discussion}
\label{sec:discussion}

We present a physics-coded neural Koopman framework for modeling parametric PDE dynamics. The method represents spatial modes and temporal evolution using neural fields conditioned on physical codes, yielding a structured latent representation in which space, time, and physics are parameterized separately. This formulation supports long-horizon rollouts and interpolation across unseen physical configurations.

Unlike black-box sequence models that directly regress future states, our approach exposes a linear evolution in a learned latent space. This structure accounts for the improved stability and efficiency observed in our experiments, as well as robust generalization across variations in physical parameters, boundary conditions, and geometry. Ablation results further confirm that enforcing explicit spectral structure and long-horizon consistency is essential for reliable performance.
\section{Conclusion}
\label{sec:conclusion}

We introduced a physics-coded neural Koopman framework for learning parametric PDE dynamics with explicit spectral structure, where physics-conditioned neural fields parameterize spatial modes and temporal evolution to enable stable long-term rollout, efficient inference, and generalization across varying physical parameters, boundary conditions, and geometries, while achieving the lowest prediction error with near-minimal memory and inference cost across all benchmarks.

These results indicate that learning parameter-conditioned Koopman representations provides a viable alternative to black-box autoregressive models for parametric PDEs. More broadly, this work underscores the benefit of incorporating explicit dynamical structure into neural models to improve reliability and interpretability in scientific machine learning.
\section{Impact Statement}
This work advances the field of scientific machine learning. A primary impact of our method is improved interpretability for learned dynamics. By constructing a fully transparent, linear model via a neural representation, we provide a counterpoint to black-box neural operators, which can aid in scientific validation and discovery.

Furthermore, our approach has a positive computational impact. Long-term simulation is reduced to efficient matrix multiplications, eliminating the need for repeated, energy-intensive neural network evaluations. This significantly lowers the computational cost and energy footprint of simulations, making them more accessible for deployment on standard hardware, including CPU-only systems. We foresee no unique negative societal consequences beyond those typical of the field.

\bibliography{references}
\bibliographystyle{icml2026}

\newpage
\appendix
\onecolumn
\section{Implementation Details}
\begin{table*}[!htbp]
\centering
\caption{Network architectures and loss configurations across experiments.}
\label{tab:architecture}
\renewcommand{\arraystretch}{1.15}

\begin{tabular}{c c c c c c}
\toprule
\textbf{Experiments} &
\textbf{$\phi$ MLP} &
\textbf{$\lambda$ MLP} &
$\alpha$ &
$\beta$ &
\textbf{\# Modes}\\
\midrule
\textbf{Burgers' Equation} &
$64 \times 3$ (Sine) &
$64 \times 3$ (ELU) &
$0.90$ &
$0.10$ &
$2$\\

\textbf{Double Shear} &
$128 \times 3$ (Sine) &
$64 \times 3$ (ELU) &
$0.05$ &
$0.95$ &
$6$\\

\textbf{K\'arm\'an Vortex Street} &
$128 \times 3$ (Sine) &
$64 \times 3$ (ELU) &
$0.10$ &
$0.90$ &
$5$\\

\textbf{Airfoil} &
$128 \times 3$ (Sine) &
$64 \times 3$ (ELU) &
$0.05$ &
$0.95$ &
$5$\\
\bottomrule
\end{tabular}
\end{table*}

\subsection{Pseudoinverse}
\refeq{eq:pinv} requires the computation of a pseudoinverse.
In practice, differentiating through a pseudoinverse in automatic differentiation frameworks incurs substantial memory overhead, since large intermediate tensors must be retained for gradient computation. We therefore implement this operation as an equivalent least-squares solve, achieving the same numerical result while avoiding the high memory cost of backpropagating through the pseudoinverse.

\subsection{Optimization}
All networks are trained using the \texttt{Adam} optimizer.
For the $\phi$ networks, we use a learning rate of $1\times10^{-4}$ with
$(\beta_1,\beta_2)=(0.9,0.99)$, $\epsilon=10^{-15}$, and $L_2$ weight decay
$1\times10^{-6}$.
A \texttt{cosine annealing} learning rate scheduler is applied with a minimum learning
rate of $1\times10^{-6}$.
For the $\lambda$ networks, \texttt{Adam} is used with learning rates in
$\{10^{-5},10^{-3}\}$ depending on the experiment, while keeping all other
optimizer parameters identical.
All experiments are trained with batch size $16$ for up to $500$ epochs.

\subsection{Architecture Justification}
\begin{figure}[t]
    \centering
    \includegraphics[width=0.8\linewidth]{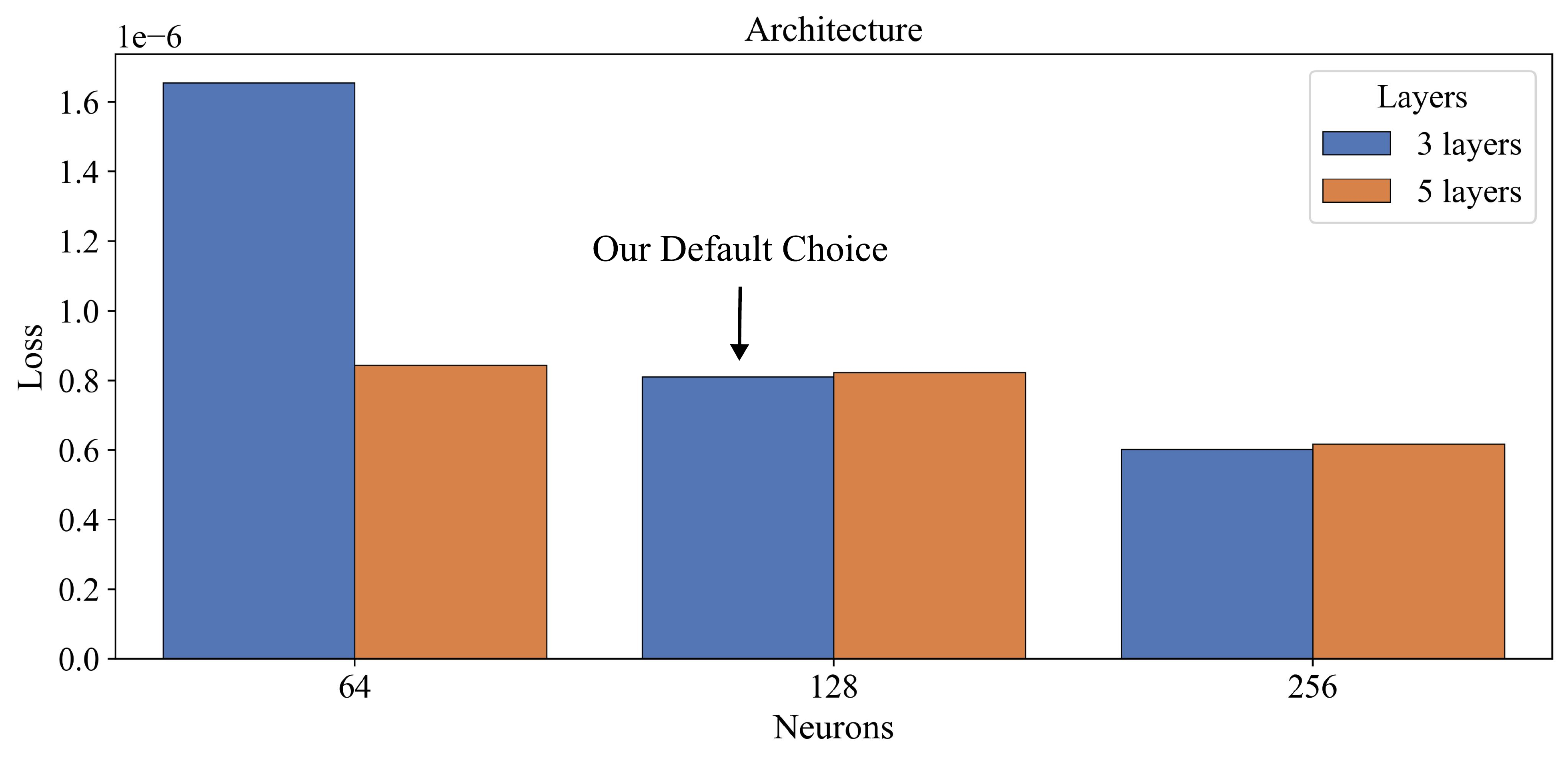}
    \caption{\textbf{Architecture ablation on network width and depth.} We evaluate the impact of the number of neurons per layer (64, 128, 256) and network depth (3 vs. 5 layers) on the training loss. While increasing model capacity generally reduces the loss, the improvement saturates beyond 128 neurons. Deeper networks do not consistently outperform shallower ones under the same width. In our experiments, we choose the 3-layer configuration with 128 neurons as our default setting.}
    \label{fig:ablation_architecture}
\end{figure}

We test different neuron counts and layer depths, and select 128 neurons with 3 layers as the default, balancing model capacity against overall size (\reffig{fig:ablation_architecture}).

\subsection{Baselines}
\label{sec:baselines}

We compare against six representative baselines: KAE, FNO, KNO, P-DMD, PDE-T, and RKN. 
For fair comparison, all baselines are trained and evaluated on the same train/test splits and rollout protocols, and we report metrics computed over the entire trajectory.

\paragraph{KAE.}
We follow the optimization and training settings reported in KAE~\cite{azencot2020forecasting}: We use \texttt{Adam} with \texttt{PyTorch} default hyperparameters, apply gradient clipping at $0.05$, then train for 2000 epochs with batch size 20 using step-based rollout training (steps $=8$, steps\_back $=8$, backward $=1$) with pre-conditioning enabled. We adopt a step-decay learning-rate schedule with decay factor $0.2$; since our dataset differs from that used in KAE, we extend the learning-rate update milestones to ${100, 300, 600, 1000, 1500}$ to ensure stable convergence.

\paragraph{FNO.}
We follow the optimization and training settings reported in FNO~\cite{li2020fourier}: We use two settings depending on the data dimensionality. \textbf{1D FNO:} we set the Fourier rank to $16$, use \texttt{Adam} with learning rate $10^{-3}$, apply $L_2$ weight decay of $1\times10^{-4}$, and train with batch size $B=20$; all other optimizer hyperparameters follow \texttt{PyTorch}'s default settings. We use a StepLR scheduler (\texttt{torch.optim.lr\_scheduler.StepLR}) with step\_size$=100$ and $\gamma=0.5$, and train for 500 epochs. \textbf{2D FNO:} we set the Fourier rank to $12$ along each spatial dimension and keep the remaining settings identical.

\paragraph{KNO.}
We follow the optimization and training settings reported in KNO~\cite{xiong2024koopman}: We use \texttt{Adam} with \texttt{PyTorch} default hyperparameters and a StepLR schedule with step\_size$=100$ and $\gamma=0.5$, training for 500 epochs. For \textbf{1D} dataset, we set rank$=16$, operator\_size$=16$, and operator\_power$=8$; for \textbf{2D} dataset, we set rank$=16$, operator\_size$=32$, and operator\_power$=8$. It is worth noting that KNO cannot handle sequences of fields with spatial dimension greater than 1; therefore, KNO is not applicable to the K\'arm\'an and airfoil datasets.

\paragraph{P-DMD.}
We follow the optimization and training settings reported in P-DMD~\cite{pDMD}: We use the same number of basis functions as in our method for each experiment. Note that one mode corresponds to two basis functions (a conjugate pair); accordingly, for P-DMD we use $4$ basis functions for Burgers' equation, $12$ for Double Shear, $10$ for K\'arm\'an vortex street, and $10$ for the airfoil benchmark.

\paragraph{PDE-T.}
We follow the optimization and training settings reported in PDE-T~\cite{holzschuh2025pde}: We keep all hyperparameters consistent with the default configuration (batch size$=8$, base learning rate$=4\times10^{-5}$, patch size$=4$, in\_channels$=2$, out\_channels$=2$, and sample size$=128$). Note that PDE-T does not support 1D datasets, and thus we do not report results on Burgers' equation.

\paragraph{RKN.}
We follow the optimization and training settings reported in RKN~\cite{xu2025reskoopnetlearningkoopmanrepresentations}: We use \texttt{Adam} with lr $=10^{-5}$, with all other optimizer hyperparameters set to \texttt{PyTorch} defaults. We set the dictionary size to $300$ and train for $200$ steps with batch size $256$. We use a custom learning-rate scheduler that decays the current learning rate by a factor of $0.8$ whenever the training loss increases between consecutive steps. Notably, RKN is highly memory-intensive: for the K\'arm\'an vortex street and airfoil benchmarks, it runs out of GPU memory even after reducing the batch size and model size, so we do not report RKN results on these two datasets.

\subsection{Burgers Equation}
\label{app:burgers}
We solve the 1D viscous Burgers equation on the periodic domain $x\in[0,1), t\in[0,1]$:
\begin{equation}
\label{eq:burgers_pde}
\partial_t u + u\,\partial_x u = \nu\,\partial_{xx}u,
\qquad u(x,0) = u_0(x),
\qquad u(0,t) = u(1,t),
\qquad \partial_x u(0,t) = \partial_x u(1,t).
\end{equation}

The initial condition $u_0$ is a periodic Gaussian random field (GRF):
\begin{equation}
\label{eq:burgers_ic}
u_0(x)
= m + \sum_{k=1}^{K}\Big(\alpha_k \cos\!\big(2\pi k (x-\tfrac12)\big)
+ \beta_k \sin\!\big(2\pi k (x-\tfrac12)\big)\Big),
\end{equation}
with $\alpha_k,\beta_k\sim\mathcal{N}(0,\lambda_k^2)$ and $\lambda_k = \sqrt{2}\,|\sigma|\Big((2\pi k)^2+\tau^2\Big)^{-\gamma/2}.$

In our generator, we set $m=0$ and use $K = s/2$ with $s=1024$ grid points. We fix $\gamma=2.5$, $\tau=7.0$, $\sigma=49.0$, and simulate to $T=1.0$ with 200 time steps using a Fourier pseudo-spectral solver with 2/3 de-aliasing and an ETDRK4 time integrator. The viscosity $\nu$ is the code parameter, sampled uniformly from $0.1$ to $0.001$ (19 codes).

\begin{figure}[!t]
  \centering
  \includegraphics[width=\linewidth]{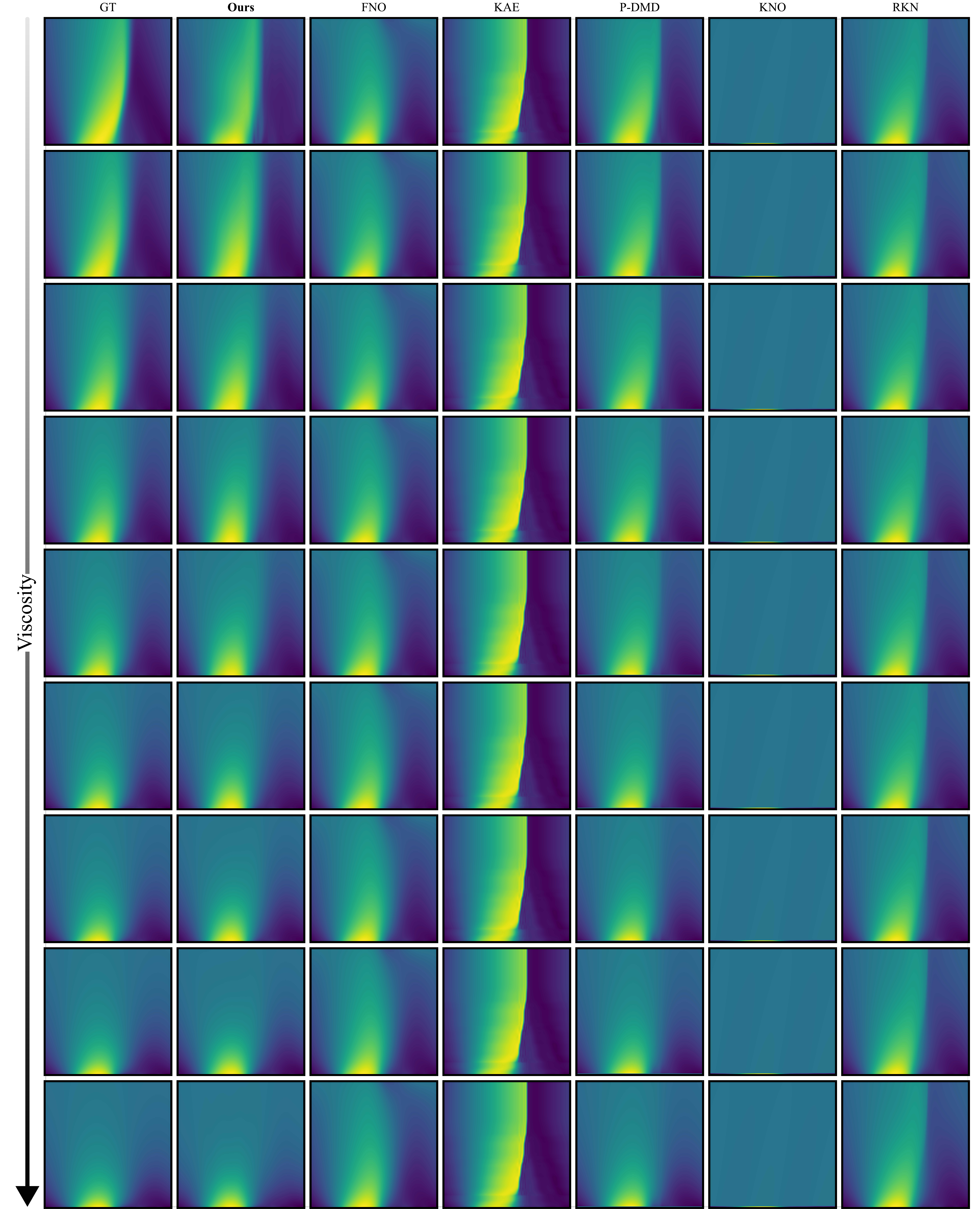}
  \caption{\textbf{Burgers' Equation across \emph{testing set} viscosities.}
  Spatiotemporal solutions $u(x,t)$ are shown as heatmaps. Each row corresponds to a different
  viscosity value \emph{from the testing set} (increasing from top to bottom, as indicated on
  the left). Columns compare the ground truth (GT) with predictions from \emph{Ours}, FNO, KAE,
  P-DMD, KNO, and RKN. All panels use the same color scale.}
  \label{fig:burgers_full_physics}
\end{figure}

\subsection{Double Shear Layer}
\label{app:double}
We consider the 2D incompressible Navier--Stokes equations in \emph{vorticity form}. Let
$\omega(x,z,t)$ denote the scalar vorticity and $\bm{u}(x,z,t)=(u(x,z,t),w(x,z,t))$ the velocity
field. The PDE is initialized by prescribing the vorticity field at $t=0$,
\begin{equation}
\omega(x,z,0)=\omega_0(x,z), \qquad (x,z)\in(0,L_x)\times(0,L_z),
\end{equation}
where $\omega_0=\nabla\times \bm{u}(\cdot,0)=\partial_x w(\cdot,0)-\partial_z u(\cdot,0)$ is
computed from the velocity initial condition in \refeq{eq:double_shear_ic}. We use a \emph{double shear layer} initial condition, parameterized by the shear separation $s\in[0.2,0.4]$, defined via the velocity field
\begin{equation}
\label{eq:double_shear_ic}
\begin{cases}
\begin{aligned}
u(x,z,0)
&= u_{\max}\!\left[\tanh\!\left(\dfrac{z - z_1}{\delta}\right)
- \tanh\!\left(\dfrac{z - z_2}{\delta}\right) - 1\right] \\
&\quad + \varepsilon\,\sin\!\left(\dfrac{4\pi x}{L_x}\right)
\left[\exp\!\left(-\left(\dfrac{z-z_1}{\delta}\right)^2\right)
+\exp\!\left(-\left(\dfrac{z-z_2}{\delta}\right)^2\right)\right],
\end{aligned}\\[4pt]
w(x,z,0) = \varepsilon\,\xi(x,z),
\end{cases}
\quad (x,z)\in(0,L_x)\times(0,L_z),
\end{equation}
where $z_1=(L_z-s)/2$ and $z_2=(L_z+s)/2$, $u_{\max}=1$, and $\epsilon=0.01$, $\delta=0.05$, and $\xi(x,z)$ is standard Gaussian noise sampled i.i.d. over the grid. so that the two layers are separated by a distance $s$. In our dataset, the separation $s$ is the code parameter: we uniformly sample $21$ values of $s$ over
$[0.2,0.4]$. We use a rectangular domain of size $L_x=2$ and $L_z=1$, discretized on a uniform grid of size $128\times 64$. The corresponding initial vorticity $\omega_0=\nabla\times \bm{u}(\cdot,0)$ consists of two oppositely signed shear layers centered at $z_1$ and $z_2$, with a small sinusoidal perturbation localized near each layer and additional Gaussian noise in the transverse velocity component. This setup triggers Kelvin--Helmholtz roll-up and subsequent vortex merging dynamics. We simulate each trajectory over $t\in[0.0,10.0]$. Since the solution changes only mildly during the initial transient ($t<2.0$), we train on the window $t\in[2.0,10.0]$, from which we uniformly sample $160$ snapshots per trajectory.

\begin{figure}[t]
  \centering
  \includegraphics[width=\linewidth]{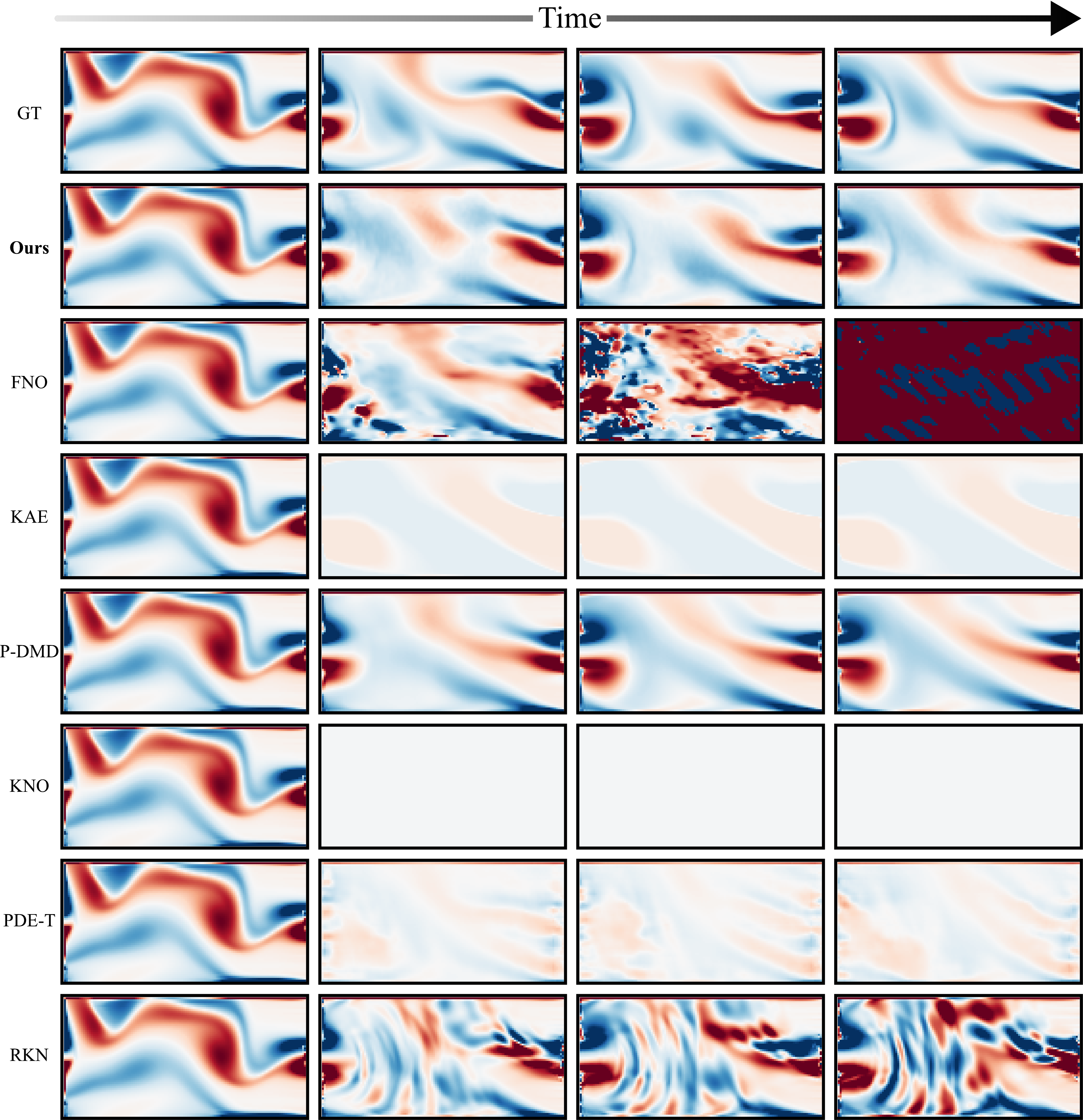}
  \caption{\textbf{Double Shear Layer Over Time.} Snapshots of the solution field at four time instants (left to right, increasing time as indicated by the arrow) with shear separation $= 0.21$. Rows show the ground truth (GT) and predictions from \emph{Ours}, FNO, KAE, P-DMD, KNO, PDE-T, and RKN.
  All panels share the same color scale.}
  \label{fig:double_sheer_full_time}
\end{figure}

\subsection{K\'arm\'an Vortex Street}
\label{app:karman}
We generate a two-dimensional cylinder wake dataset using the lattice Boltzmann method (LBM)~\cite{Lattice1998}, where the cylinder induces vortex shedding. 

The computational domain is discretized on a uniform lattice of resolution $201\times 51$, corresponding to a physical domain of length $1\,\mathrm{m}$ in the $x$ direction. All fields are defined on lattice nodes $(i,j)$, where $i\in\{0,\ldots,200\}$ and $j\in\{0,\ldots,50\}$. The lattice spacing is uniform in both directions.

The \emph{code parameter} is the center location of a circular cylinder, $c=(c_x,c_y)$, specified in lattice coordinates. We fix $c_y=25$ and vary $c_x\in\{70,71,\ldots,90\}$, uniformly sampling $21$ values. The cylinder radius is fixed to $R=5$ in lattice units. Varying $c_x$ translates the obstacle within the fixed lattice domain and alters the wake development, producing
distinct vortex-shedding trajectories in the generated LBM simulations.

All simulations are performed with inflow velocity $\bm{u}=(0.1,\,0.0)$ and kinematic viscosity $\nu=0.01$. We impose a Dirichlet inlet at the left boundary, a zero-gradient outflow at the right boundary, and no-slip walls at the top and bottom boundaries. Each simulation is run for 200 frames.

\begin{figure}[t]
  \centering
  \includegraphics[width=\linewidth]{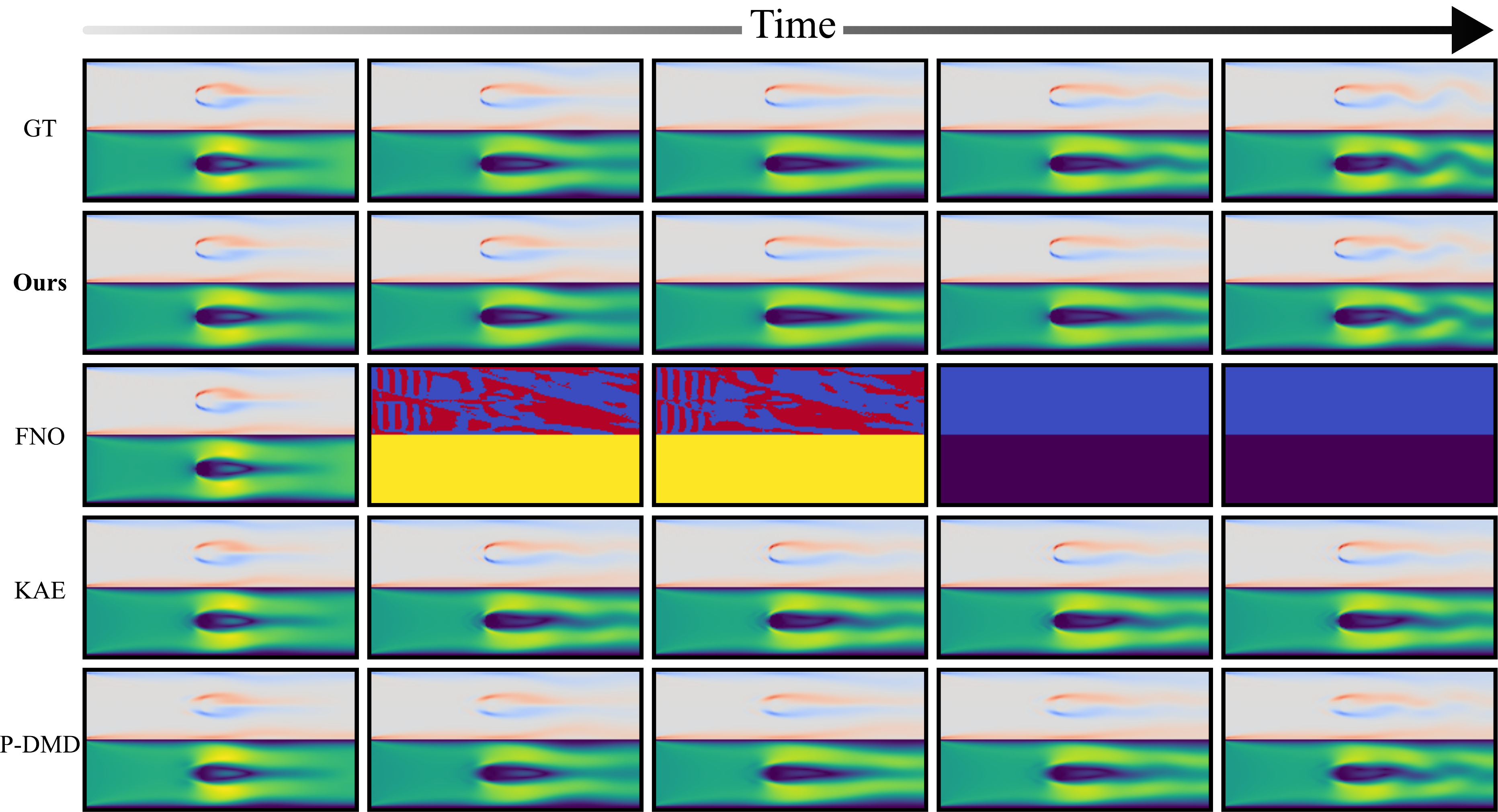}
  \caption{\textbf{K\'arm\'an Vortex Street Over Time.} Snapshots of the solution field at five time instants (left to right, increasing time as indicated by the arrow) with the cylinder placed at $x=0.435$. Rows show the ground truth (GT) and predictions from Ours, FNO, KAE, and P-DMD. All panels share the same color scale.}
  \label{fig:karman_full_time}
\end{figure}

\subsection{Airfoil}
\label{app:airfoil}
We build an airfoil-flow dataset using the same lattice Boltzmann simulation setup as in the \refsec{app:karman}, but with a uniform Cartesian lattice of size $256 \times 128$.

The airfoil boundary is parameterized by a \emph{six-parameter} Class--Shape Transformation (CST)
model with parameter vector
\[
[A_{u0},\,A_{u1},\,A_{l0},\,A_{l1},\,t_e,\,\theta_{\mathrm{cw}}],
\]
where $A_{u0}$ and $A_{u1}$ control the upper-surface shape, $A_{l0}$ and $A_{l1}$ control the
lower-surface shape, $t_e$ specifies the trailing-edge thickness, and $\theta_{\mathrm{cw}}$ is a clockwise
rotation angle about the fixed center $(0.5,0.0)$.

We define the class function
\[
C(x)=x^{N_1}(1-x)^{N_2},\qquad (N_1,N_2)=(0.5,1.0),
\]
and sample $x\in[0,1]$ using cosine spacing $x=\tfrac{1}{2}(1-\cos\beta)$ with $\beta\in[0,\pi]$.
Using a first-order Bernstein basis $B_0(x)=1-x$ and $B_1(x)=x$, we form the shape functions
\[
S_u(x)=A_{u0}B_0(x)+A_{u1}B_1(x),\qquad
S_l(x)=A_{l0}B_0(x)+A_{l1}B_1(x).
\]
The upper and lower surfaces are then given by
\[
z_u(x)=C(x)S_u(x)+\tfrac{1}{2}t_e x,\qquad
z_l(x)=C(x)S_l(x)-\tfrac{1}{2}t_e x.
\]
The resulting point set is assembled into a closed airfoil curve and finally rotated clockwise
by $\theta_{\mathrm{cw}}$ about $(0.5,0.0)$.

In our parameter sweep, we vary $A_{u0}\in[0.40,0.50]$, $A_{l0}\in[-0.20,-0.10]$, and
$\theta_{\mathrm{cw}}\in[0^\circ,10^\circ]$ within the specified ranges, while fixing
$A_{u1}=0.08$, $A_{l1}=-0.02$, and $t_e=0.002$.
We construct the training set by sampling 27 codes uniformly over the specified parameter ranges, and sample an additional 16 codes at random from the same ranges for testing, resulting in 43 airfoil codes in total. Each simulation is run for 200 frames.

\section{Koopman Operator and the DMD Operator}
\label{app:koopman_operator}

This supplemental section introduces the \textbf{Koopman operator} and \textbf{Dynamic Mode Decomposition (DMD)}, and explains how the \textbf{DMD operator} can be interpreted as a finite-dimensional approximation of Koopman spectral objects from data.

\subsection{Nonlinear dynamics and the observable viewpoint}

Consider a continuous-time system
\begin{equation}
\dot{x}=f(x), \quad x(t)\in\mathbb{R}^n,
\end{equation}
or an equivalent discrete-time dynamical system
\begin{equation}
x_{k+1}=F(x_k), \quad x_k\in\mathbb{R}^n.
\end{equation}
Koopman theory shifts perspective from the nonlinear evolution of the \emph{state} $x$ to the linear evolution of \emph{observables} (functions of the state). Without loss of generality, let $g:\mathbb{R}^n\to\mathbb{C}^m$ be a vector-valued observable.

\subsection{The Koopman operator: definition and key properties}

\paragraph{(1) Discrete-time Koopman operator.}
The Koopman operator $\mathcal{K}$ acts on observables $g$ as:
\begin{equation}
(\mathcal{K}g)(x)=g(F(x)).
\end{equation}
In words, $\mathcal{K}$ propagates the observable $g$ one timestep, being an equivalent operation as evolving the state $x$ through $F$ and evaluating the observable at the new state.

\paragraph{(2) Linearity.}
Crucially, even if $F$ is nonlinear, $\mathcal{K}$ is \emph{linear}:
\begin{equation}
\mathcal{K}(a g_1+b g_2)=a\,\mathcal{K}g_1+b\,\mathcal{K}g_2,
\end{equation}
because $\mathcal{K}$ acts on a space of functions (observables), not directly on states.

\paragraph{(3) Koopman eigenfunctions, eigenvalues, and modes.}
Let $\varphi_j$ be an eigenfunction with eigenvalue $\lambda_j$. Then
\begin{equation}
\mathcal{K}\varphi_j = \lambda_j \varphi_j,
\end{equation}
and along trajectories $x_k$,
\begin{equation}
\varphi_j(x_k) = \lambda_j^k\,\varphi_j(x_0).
\end{equation}
For a vector-valued observable $g(x)\in\mathbb{C}^m$, under suitable assumptions one can express
\begin{equation}
g(x_k)=\sum_{j=1}^{\infty}\phi_j\,\varphi_j(x_0)\,\lambda_j^{k},
\end{equation}
where $\phi_j\in\mathbb{C}^m$ are the \emph{Koopman modes} associated with the observable $g$.

\subsection{DMD: a data-driven linear operator for time evolution}

DMD is a data-driven method that fits a best linear map that advances measured snapshots forward in time.

\paragraph{(1) Snapshot matrices.}
Given snapshots sampled at uniform time intervals,
\begin{equation}
x_1, x_2, \dots, x_{m+1},
\end{equation}
form
\begin{equation}
X=[x_1,x_2,\dots,x_m],\qquad X'=[x_2,x_3,\dots,x_{m+1}].
\end{equation}

\paragraph{(2) The DMD operator.}
DMD seeks a matrix $A$ such that
\begin{equation}
X' \approx A X
\end{equation}
in a least-squares sense. The minimizer is
\begin{equation}
A = X' X^{+},
\end{equation}
where $X^{+}$ denotes the Moore--Penrose pseudoinverse of $X$. This $A$ is called the \emph{DMD operator}.

\paragraph{(3) Low-rank (projected) DMD.}
In practice, a rank-$r$ truncated SVD is used:
\begin{equation}
X \approx U_r\Sigma_r V_r^*.
\end{equation}
The reduced operator is
\begin{equation}
\widetilde{A}=U_r^*X'V_r\Sigma_r^{-1}.
\end{equation}
Its eigen-decomposition is given by
\begin{equation}
\widetilde{A}w_j=\mu_j w_j,
\end{equation}
where $\mu_j$ are DMD eigenvalues. One common form for the corresponding DMD modes is
\begin{equation}
\psi_j = \frac{1}{\mu_j}X'V_r\Sigma_r^{-1}w_j,
\end{equation}
(with equivalent expressions depending on the exact algorithmic variant).

If the sampling interval is $\Delta t$, discrete-time eigenvalues can be mapped to continuous-time growth/decay rates and frequencies by
\begin{equation}
\omega_j=\frac{\log(\mu_j)}{\Delta t}.
\end{equation}

\subsection{Relationship between Koopman and DMD}

The Koopman operator is generally infinite-dimensional, acting on an (infinite) function space. DMD, by contrast, produces a finite-dimensional matrix $A$ that advances data snapshots.

If the snapshots $x_k$ are treated as evaluations of some observable $g(x)$ (often $g(x)=x$, i.e., direct state measurements), and the span of the chosen observables is approximately invariant under $\mathcal{K}$, then DMD approximates the action of the Koopman operator projected onto that finite-dimensional observable subspace. In this sense, DMD provides numerical estimates of Koopman spectral quantities (eigenvalues and modes) associated with the measured observables.

More broadly, if one uses a lifted observable map $y_k=\Psi(x_k)$ (e.g., polynomial features, kernels, or neural-network embeddings), then applying the same regression/DMD procedure yields EDMD / kernel DMD, which approximates Koopman on the span of $\Psi$.




\end{document}